\documentclass[journal]{IEEEtran}
\usepackage{amsmath,amsfonts}
\usepackage{algorithmic}
\usepackage{algorithm}
\usepackage{array}
\usepackage[caption=false,font=normalsize,labelfont=sf,textfont=sf]{subfig}
\usepackage{textcomp}
\usepackage{stfloats}
\usepackage{url}
\usepackage{verbatim}
\usepackage{graphicx}
\usepackage{cite}
\usepackage{booktabs}
\usepackage{multirow}
\usepackage{amssymb}
\usepackage[table]{xcolor}
\begin{document}

\title{OneEEG: A Unified Signal-Visual Multimodal Language Model with Task-Aware Mixture of Projectors for Multi-Task EEG Understanding}

\author{Jiacheng Zhang, Xueyi Zhang, Xin Lu, Siqi Cai, Haizhou Li
\thanks{This work was supported in part by the Program for Guangdong Introducing Innovative and Entrepreneurial Teams under Grant 2023ZT10X044, Shenzhen Loop Area Institute under Grant FPF10120250012, the National Science and Technology Major Project for Brain Science and Brain-like Intelligence Technology (2025ZD0215700), and the National Natural Science Foundation of China (72421002).}
\thanks{Jiacheng Zhang and Haizhou Li are with the School of Artificial Intelligence, The Chinese University of Hong Kong, Shenzhen, China, and Language, Intelligence, and Machine Center, Shenzhen Loop Area Institute, Shenzhen, China (e-mail: jiachengzhang@slai.edu.cn; haizhouli@slai.edu.cn).}
\thanks{Siqi Cai is with the School of Intelligence Science and Engineering, Harbin Institute of Technology, Shenzhen, China, and Language, Intelligence, and Machine Center, Shenzhen Loop Area Institute, Shenzhen, China (e-mail: siqicai@slai.edu.cn).}
\thanks{Xueyi Zhang is with the School of Computing, National University of Singapore, Singapore (e-mail: xueyi.zhang@nus.edu.sg).}
\thanks{Xin Lu is with the College of Systems Engineering, National University of Defense Technology, Changsha, China (e-mail: xin.lu.lab@outlook.com).}
\thanks{Siqi Cai and Haizhou Li are co-corresponding authors.}
}

\markboth{IEEE Transactions on Multimedia}%
{Zhang \MakeLowercase{\textit{et al.}}: OneEEG for Multi-Task EEG Understanding}


\maketitle

\begin{abstract}
Electroencephalography (EEG) is widely used in clinical assessment, brain-computer interfaces, and cognitive-state analysis, yet EEG modeling remains fragmented across task paradigms. Recent multimodal large language models (MLLMs) provide a promising interface for flexible multimodal reasoning, but their ability to understand EEG inputs remains limited. 
In this work, we propose OneEEG, a unified EEG-MLLM framework for multi-task EEG understanding. By training on our curated \textbf{OneEEG-Instruct}, a multi-task EEG instruction-tuning dataset, OneEEG jointly models numerical EEG representations and visual EEG representations, and introduces a \textbf{task-aware Mixture of Projectors (MoP)} to align heterogeneous EEG features across different task families with the LLM embedding space.
We further establish \textbf{OneEEG-Bench}, a benchmark for evaluating current MLLMs and our method on multi-task EEG understanding. 
Experiments show that existing MLLMs have limited ability to interpret EEG inputs, while OneEEG achieves stronger performance by integrating numerical and visual EEG representations with task-aware EEG-to-LLM projection.
Extensive ablations and analyses further reveal the effects of visual EEG renderings, MoP variants, router behavior, and feature dispersion across EEG tasks.
 
\end{abstract}



\section{Introduction}

\IEEEPARstart{E}{lectroencephalography} (EEG) has long served as a fundamental modality for studying neural activity, owing to its non-invasive acquisition, relatively low cost, and high temporal resolution~\cite{acharya2018seizure_survey,lee2025bcireview}. These properties have made EEG central to a wide range of research and application scenarios, including neurological assessment\cite{fratangelo2025point}, seizure analysis\cite{acharya2018seizure_survey}, sleep monitoring~\cite{satapathy2025sleepreview}, brain-computer interfaces~\cite{lawhern2018eegnet}, cognitive-state estimation~\cite{yedukondalu2025cognitive}, and affective computing~\cite{wang2025emotion_reco}. Despite this broad relevance, EEG modeling remains largely fragmented across task paradigms. EEG applications are often associated with distinct experimental protocols, channel configurations, subject populations, annotation schemes, and physiological phenomena of interest. More importantly, the discriminative patterns in EEG are highly task-dependent, spanning temporal dynamics, spectral characteristics, and spatial distributions. As a result, most EEG models are still developed under task-specific assumptions~\cite{liu2024msvtnet,liu2024graph, ozdemir2021emotion_method_2}, with dedicated preprocessing procedures and model architectures tailored to individual datasets or label spaces. This task-specific modeling paradigm limits the development of EEG systems that can support heterogeneous tasks under a unified representation and interaction framework.


\begin{figure}
    \centering
    \includegraphics[width=0.9\linewidth]{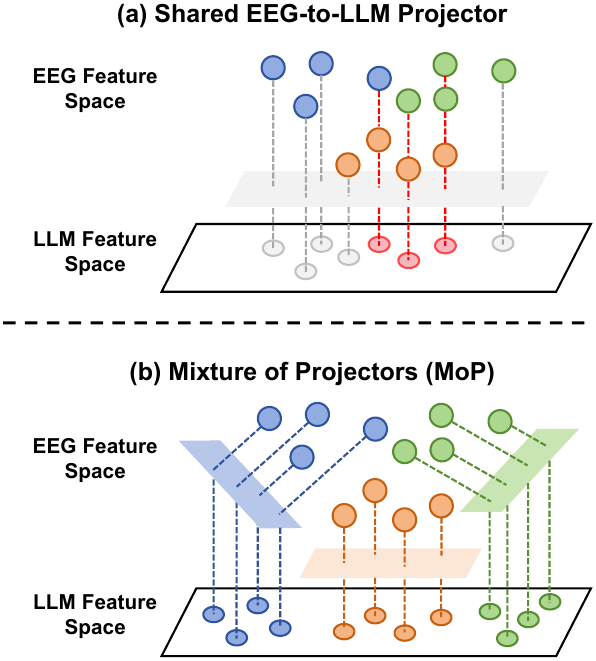}
    \caption{
    Motivation of MoP. Colored points denote EEG features from different task families or datasets. 
    A shared EEG-to-LLM projector forces heterogeneous EEG features through a single projector, which may cause cross-task conflicts as indicated by the red dashed paths. 
    MoP mitigates this bottleneck by aligning task-dependent EEG features through task-aware projection.
}
    \label{fig:intro1}
\end{figure}


Recent advances in large language models (LLMs) have motivated efforts to connect EEG signals with LLMs or multimodal LLMs (MLLMs) to enable multi-task EEG understanding ~\cite{zeng2025wavemind,gu2025cerebragloss,ma20262e2llm}. Existing efforts generally follow two directions. One line of work adopts a standard MLLM paradigm ~\cite{zeng2025wavemind, ma20262e2llm}, where the EEG encoder extracts features from numerical signals, and a shared projector maps them into the LLM embedding space. 
Another line of work studies image-form EEG understanding, where EEG signals are rendered as waveform plots, spectra, or topographic maps and interpreted by vision-language models~\cite{gu2025cerebragloss,qiu2026eegvlm}.
However, these paradigms suffer from two important limitations. \textbf{(1) Single projector.} A single shared projector may be insufficient for aligning heterogeneous EEG representations with the semantic space of an LLM. Unlike natural images or text, EEG is a low-signal-to-noise physiological signal with weak intrinsic semantic information, and EEG features can be highly dispersed across different tasks. 
Projecting such heterogeneous features through a single mapping module may introduce cross-task interference and weaken task-specific semantic alignment.
\textbf{(2) Single input form.} 
Existing approaches often support only one form of EEG input, either a frame of multiple 1-dimensional EEG signals, i.e., numerical EEG representations, or a frame of 2-dimensional visual image of EEG signals, i.e., visual EEG representations.  
Numerical EEG representations preserve the original signal form and retain fine-grained temporal information, while 
visual EEG representations capture the co-occurrence of EEG events across channels within a frame. Each has its pros and cons.
In many real-world settings, such as sleep monitoring, clinical diagnosis, and brain function assessment, visual EEG representations are useful for experts to inspect and interpret EEG recordings. 
On the other hand, although numerical signals provide more direct access to fine-grained EEG information, an encoder pre-trained on one set of numerical signals may not be compatible with others.
In contrast, visual EEG representations provide a more flexible interface for datasets with non-standard channel systems. 
These considerations motivate us to embrace both numerical and visual EEG representations.

Therefore, we propose \textbf{OneEEG}, a unified EEG-MLLM framework for multi-task EEG understanding that jointly models numerical EEG representations and visual EEG representations. 
OneEEG adopts a dual-stream architecture, in which the numerical signal stream extracts neural features from numerical EEG representations, while the visual stream processes rendered EEG waveform images. 
The features from both streams are projected into the semantic space of the LLM backbone, supporting EEG-oriented question answering over both signal-derived features and visual EEG patterns. 
To address the heterogeneous alignment problem across EEG tasks, OneEEG further introduces a task-aware \textbf{Mixture of Projectors (MoP)} module. Rather than relying on a single shared EEG-to-LLM projection, MoP employs a task-aware routing mechanism over a bank of projectors to align EEG representations from different task families with the LLM embedding space through task-aware projection. By combining dual-stream EEG input modeling with task-aware MoP, OneEEG provides a unified baseline for heterogeneous EEG understanding.

To support the training of OneEEG, we curate \textbf{OneEEG-Instruct}, a multi-task EEG instruction-tuning dataset that transforms heterogeneous EEG tasks into natural-language supervision. OneEEG-Instruct supports dual-stream EEG inputs, covering both numerical EEG representations and waveform-based visual EEG representations, and spans diverse EEG task paradigms involving clinical, event-related, cognitive, and motor-intent understanding. Beyond aggregating heterogeneous tasks, OneEEG-Instruct adopts a hierarchical instruction design to integrate both task-specific EEG comprehension and general signal-level perception. The task-QA tier is closely tied to downstream EEG tasks and captures dataset-specific prediction objectives, whereas the basic-QA tier focuses on task-agnostic EEG attributes that reflect low-level signal characteristics.
In parallel, we establish \textbf{OneEEG-Bench}, an EEG-MLLM benchmark for systematically assessing EEG understanding in multimodal language models. OneEEG-Bench evaluates general closed-source MLLMs, open-source MLLMs, medical MLLMs, EEG-specialized MLLMs, and our proposed OneEEG under a meticulously designed evaluation protocol. Together, OneEEG-Instruct and OneEEG-Bench provide a consistent basis for training EEG-specific MLLMs and evaluating the capability of existing multimodal language models in multi-task EEG understanding.


{In summary, our main contributions are as follows:
\begin{itemize}
    \item We propose \textbf{OneEEG}, a unified EEG-MLLM framework that supports both numerical EEG representations and visual EEG representations for multi-task EEG understanding. 
    We further introduce a task-aware \textbf{Mixture of Projectors (MoP)} to mitigate the single-projector bottleneck and improve task-aware EEG-to-LLM alignment across heterogeneous EEG tasks.
    \item We construct \textbf{OneEEG-Instruct} and \textbf{OneEEG-Bench} for training and evaluation, covering diverse EEG task families, channel settings, and closed-set QA formats, with hierarchical annotations spanning task-oriented EEG reasoning and basic signal attribute perception.
    \item We conduct extensive experiments on OneEEG-Bench to systematically evaluate current MLLMs and analyze the effectiveness of OneEEG across diverse EEG tasks, demonstrating the superior performance of OneEEG over existing baselines.
\end{itemize}}

\section{Related Works}
In this section, we review the literature relevant to OneEEG, including EEG representation learning and foundation models, multimodal large language models for EEG understanding, and the development of the mixture of experts mechanism.

\subsection{EEG Representation Learning and Foundation Models}
EEG modeling has traditionally been developed in task-specific settings, where models are designed for individual applications such as seizure detection \cite{acharya2018seizure_survey}, sleep staging\cite{satapathy2025sleepreview}, motor imagery decoding\cite{liu2024msvtnet}, or emotion recognition\cite{wang2025emotion_reco,jin2024pgcn,pi2026enhancing}. Such task-dependent pipelines often rely on dedicated preprocessing strategies, feature designs, and supervised objectives, making it difficult to build transferable EEG models across heterogeneous paradigms. To improve the generality of EEG representations, recent studies have explored EEG foundation models trained on large-scale or multi-dataset EEG corpora. Representative works, including LaBraM~\cite{jiang2024labram}, EEGPT~\cite{wang2024eegpt}, Brant~\cite{zhang2023brant}, CBraMod~\cite{wang2025cbramod}, BrainOmni~\cite{xiao2026brainomni}, BIOT~\cite{yang2023biot}, Neuro-GPT~\cite{cui2024neurogpt}, and NeuroLM~\cite{jiang2024neurolm}, adopt self-supervised or large-scale pretraining strategies to learn transferable representations from EEG signals.

These models share the goal of moving beyond isolated task-specific training toward general-purpose EEG encoders. 
Although EEG foundation models are primarily designed as representation encoders and are not directly equipped with natural-language interaction or instruction-following capabilities, they have substantially advanced transferable EEG representation learning and provide a strong foundation for building more general EEG physiological signal understanding systems. 
Our work builds on this progress by integrating EEG signal representations with multimodal language models for unified, instruction-driven EEG understanding.

\subsection{Multimodal Large Language Models for EEG Understanding}
Recent advances in multimodal large language models (MLLMs) have enabled strong visual-language reasoning across natural images, documents, charts, and medical images. General-purpose MLLMs, such as GPT-5.5~\cite{gpt55}, Gemini-3 Pro~\cite{gemini3pro}, Qwen VL series~\cite{bai2025qwen25vl,bai2025qwen3vl}, and LLaVA series~\cite{an2026llavaonevision2}, have demonstrated impressive open-ended visual understanding capabilities, while medical MLLMs, such as Lingshu~\cite{xu2025lingshu} and MedGemma~\cite{sellergren2026medgemma}, have extended multimodal reasoning to biomedical and clinical scenarios. However, visual EEG representations differ substantially from both natural images and conventional medical images, since waveforms, spectra, and topographic maps are visual renderings of physiological time-series signals. Their interpretation depends on temporal rhythms, frequency-band patterns, and task-specific neural phenomena. Therefore, despite the rapid progress of MLLMs, their capability for EEG understanding remains insufficiently studied, motivating systematic evaluation across heterogeneous EEG tasks.

Recent studies have begun to explore EEG-oriented LLMs and MLLMs. CerebraGloss~\cite{gu2025cerebragloss} constructs an instruction-tuned vision-language model for fine-grained clinical EEG interpretation based on waveform images and structured instruction data. WaveMind~\cite{zeng2025wavemind} aligns EEG signals with textual and visual modalities to build a conversational EEG model. EEG-VLM~\cite{qiu2026eegvlm} investigates EEG image-based vision-language modeling for sleep stage prediction through hierarchical visual feature alignment and language-guided reasoning. These works demonstrate the potential of language-model interfaces for EEG analysis. However, most existing efforts still emphasize either signal-to-language modeling or visual EEG interpretation alone, while unified support for both numerical EEG representations and visual EEG representations remains underexplored. 
In addition, existing models, including general MLLMs, medical MLLMs, and emerging EEG-oriented LLM/MLLM methods, have not been comprehensively evaluated under a unified benchmark across diverse EEG task paradigms. These gaps motivate unified signal-visual EEG modeling and systematic benchmarking for EEG understanding.

\subsection{Mixture-of-Experts for EEG Analysis}
Mixture-of-Experts (MoE) architectures have been widely studied as a mechanism for conditional computation and expert specialization~\cite{wang2022adamix, chen2024llavamole, zoph2022stmoe, zhang2021enhancing}. By routing inputs to different experts, MoE-based models can allocate specialized capacity to heterogeneous data distributions while avoiding the need for a single shared module to fit all patterns. This idea is particularly relevant to EEG modeling, where heterogeneity arises from multiple sources, including tasks, subjects, channels, frequency patterns, and recording protocols. Recent EEG studies have explored MoE designs from different perspectives. 
EEGMamba~\cite{gui2024eegmamba} introduces task-aware MoE to enable coverage of multiple EEG tasks in Mamba.
BrainMoE~\cite{mabrainmoe} employs channel-wise expert routing to model channel-dependent neural information, while EEGMoE~\cite{gao2026eegmoe} explores domain-decoupled experts for self-supervised EEG representation learning. 

Despite these advances, existing EEG-MoE methods are primarily concerned with EEG feature extraction and representation learning. In the context of EEG-MLLMs, the key challenge further involves modality alignment: heterogeneous EEG features with weak intrinsic semantic information must be transformed into the LLM feature space. This makes the EEG-to-LLM projection stage a critical bottleneck. In contrast to encoder-level or classifier-level MoE designs, our proposed MoP introduces task-aware specialization at the projection stage, guiding different EEG task families to specialized EEG-to-LLM projections. This provides a lightweight mechanism for mitigating cross-task alignment conflicts in instruction-following EEG understanding.

\section{Data Construction}

\begin{figure}[t]
    \centering
    \includegraphics[width=0.8\linewidth]{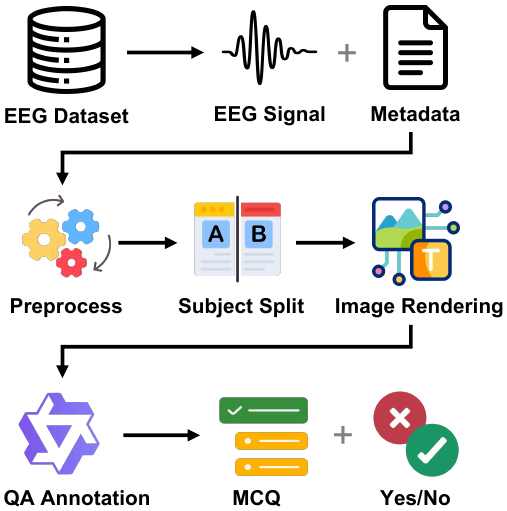}
    \caption{
    Construction pipeline of OneEEG-Instruct and OneEEG-Bench.
    Raw EEG datasets are first converted into EEG signal frames with metadata, followed by preprocessing, subject-level splitting, and EEG image rendering. 
    Question-answer annotations are then generated from the processed signals, images, and metadata, forming label-constrained MCQ and yes/no QA samples.
    }
    \label{fig:data_construction}
\end{figure}

We now discuss how we construct OneEEG-Instruct and OneEEG-Bench. 
We first demonstrate the source datasets and task taxonomy, then describe the process of EEG preprocessing and visual rendering, and finally explain how label-constrained question-answer samples are generated.
Fig.~\ref{fig:data_construction} provides an overview of the dataset construction pipeline.
(The EEG datasets used in this study are either publicly available or accessed through approved data-use requests. 
No new human-subject data were collected by the authors, and informed consent and ethical approvals were handled by the original dataset providers.)

\subsection{Dataset Sources and Task Taxonomy}
\label{taxonomy}

OneEEG-Instruct and OneEEG-Bench are built from multiple public EEG datasets covering diverse task scenarios and channel configurations. 
As summarized in Table~\ref{tab:dataset_composition}, we organize these datasets into six task families: motor, state, clinical, event, sleep, and vigilance. 
Motor and state datasets focus on movement-related and cognitive/affective EEG patterns, clinical datasets involve neurological diagnosis, event datasets contain transient EEG events such as artifacts, epileptiform discharges, and seizures, while sleep and vigilance datasets aim at sleep staging and drowsiness estimation.

We further divide the datasets according to their compatibility with a standard 19-channel 10--20 setting. 
Specifically, we define a common 19-channel subset from the international 10--20 system\cite{klem19991020channel} (Fp1, Fp2, F7, F3, Fz, F4, F8, T7, C3, Cz, C4, T8, P7, P3, Pz, P4, P8, O1, O2). 
This subset covers major scalp regions, including frontal, central, parietal, occipital, temporal, and midline areas, and provides a compact channel system that can be consistently shared across multiple EEG datasets. 
Datasets that can be reliably mapped to this subset are categorized into the standard setting.
Datasets that cannot be reliably mapped to the standard 19-channel 10--20 setting are categorized into the non-standard setting. 
The non-standard setting includes recordings with bipolar derivations, recordings concentrated on specific functional regions, and cases where the available channels do not match the selected 19-channel subset.

\subsection{EEG Preprocessing and Visual Rendering}



For each source dataset, we parse the raw EEG recordings together with the official annotations and convert them into a unified frame-level format. 
The EEG signals are processed with basic signal preprocessing, including band-pass filtering, notch filtering, resampling, and dataset-specific channel handling. 
Datasets in the standard setting are mapped to the standard 19-channel 10--20 subset defined above, while datasets in the non-standard setting keep their native channel configurations. 
Most datasets are segmented into 4-second frames, while SleepEDF follows the standard 30-second sleep epoch. 
Each frame is associated with the corresponding task label according to the original dataset annotation and saved as a numerical EEG representation frame for signal-side modeling when applicable.

{For each EEG frame, we also render the multi-channel frame into a time-channel 2D waveform image. In this way, we form a signal-image pair between the EEG signal frame and its visual image, providing two views for one EEG frame as a numerical time series or as a visual EEG representation.}

We further split the data into OneEEG-Instruct and OneEEG-Bench at the subject level when subject identifiers are available, enabling evaluation of cross-subject generalization rather than memorization of subject-specific patterns. We also generate spectrogram images as an auxiliary visual representation for analysis, but the main training and evaluation protocol is based on waveform images.

\begin{table}[t]
\centering
\caption{
Dataset composition of OneEEG-Instruct and OneEEG-Bench.
}
\label{tab:dataset_composition}
\resizebox{0.92\linewidth}{!}{%
\begin{tabular}{@{}llccc@{}}
\toprule
\textbf{Dataset} & \textbf{Family} & \textbf{Setting} & \textbf{Train/Eval} & \textbf{Format} \\
\midrule
BCIC-IV-2a \cite{brunner2008bciciv2a} & Motor & Std. & 4.6k / 320 & MCQ+Y/N \\
EEG Motor \cite{eegmotor} & Motor & Std. & 17.7k / 320 & MCQ+Y/N \\
\midrule
EEGMAT \cite{zyma2019eegmat} & State & Std. & 3.8k / 240 & MCQ+Y/N \\
SEED-IV \cite{seediv} & State & Std. & 60.1k / 360 & MCQ+Y/N \\
\midrule
TUAB ~\cite{tuh_web, obeid2016tuh} & Clinical & Std. & 50.0k / 300 & Y/N \\
TUEP ~\cite{tuh_web, obeid2016tuh} & Clinical & Std. & 50.9k / 300 & Y/N \\
\midrule
TUAR ~\cite{tuh_web, obeid2016tuh, hamid2020tuar} & Event & Std. & 65.4k / 240 & MCQ+Y/N \\
TUEV ~\cite{tuh_web, obeid2016tuh} & Event & Std. & 3.9k / 360 & MCQ+Y/N \\
TUSZ ~\cite{tuh_web, obeid2016tuh, shah2018tusz} & Event & Std. & 49.0k / 320 & MCQ+Y/N \\
\midrule
SleepEDF \cite{kemp2000sleepedf} & Sleep & Non-std. & 14.0k / 270 & MCQ+Y/N \\
DREAMS \cite{dreams} & Sleep & Non-std. & 9.0k / 304 & Y/N \\
\midrule
SEED-VIG \cite{seedvig} & Vigilance & Non-std. & 74.3k / 185 & Y/N \\
\midrule
Basic attributes & Basic & Mixed & 76.4k / 609 & MCQ \\
\bottomrule
\end{tabular}%
}
\end{table}

\subsection{Question-Answer Annotation}

{We design both OneEEG-Instruct and OneEEG-Bench under a closed-set QA protocol. 
Each question is associated with a finite candidate label set, and the answer is required to be selected from this set. 
In practice, we mainly use two QA formats: multiple-choice questions and binary yes/no questions, with details summarized in Table~\ref{tab:dataset_composition}.}

\begin{figure}[t]
    \centering
    \includegraphics[width=0.9\linewidth]{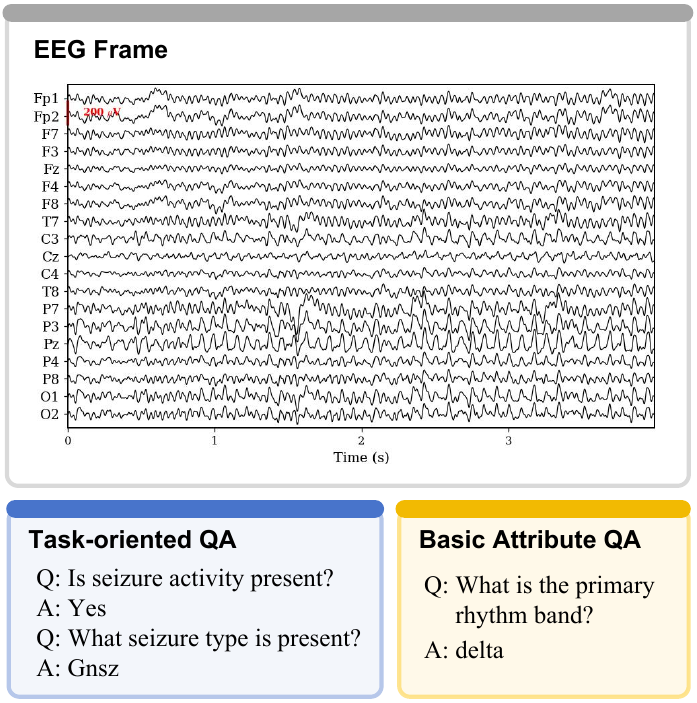}
    \caption{
    Example of the closed-set QA annotation design in OneEEG. Task-oriented QA focuses on dataset-specific EEG tasks, while basic attribute QA focuses on general signal-level properties.
    }
    \label{fig:example}
\end{figure}

To cover both dataset-specific reasoning and general signal perception, we organize QA samples into two levels, with examples shown in Fig.~\ref{fig:example}.
Task-oriented QA follows the original prediction objective of each source dataset, evaluating whether the model can perform EEG reasoning under dataset-specific label spaces.
In addition, we construct basic EEG attribute QA from subsets of the source datasets. 
These samples focus on task-agnostic signal properties, such as amplitude level, dominant frequency band, and slow-versus-fast activity tendency, enabling the evaluation of low-level EEG characteristics beyond dataset-specific task labels.

For each processed EEG frame, we construct QA annotations from its structured metadata. 
For task-oriented samples, the ground-truth answer is derived from the original dataset label or event annotation and mapped to the corresponding candidate label space. 
For basic EEG attribute samples, the answer label is computed from signal-level properties of the sampled EEG frame and converted into predefined closed-set attribute categories.

We then use Qwen3\cite{yang2025qwen3} to diversify the wording of questions under the predefined candidate label space. Specifically, we design multiple templates to facilitate the LLM to generate questions with various expressions, thereby improving the ability of OneEEG to generate accurate responses even when encountering customized queries.
During this process, the candidate labels and ground-truth answers remain unchanged, which preserves label correctness while increasing instruction diversity. 
This label-driven annotation strategy provides natural-language QA samples with deterministic closed-set answers for both training and evaluation.

\section{Method}
OneEEG is a unified EEG-oriented MLLM trained on numerical EEG signals, waveform images, and text instructions. We first present the overall architecture and the dual-stream EEG input design, then the task-aware mixture of projectors to align the EEG features with the language model space. Finally, we detail the two-stage training strategy used for projector alignment and multimodal instruction tuning.

\subsection{Overview of OneEEG}

Fig.~\ref{fig:framework} illustrates the overall architecture of OneEEG. 
Given an EEG frame and a text instruction, OneEEG constructs a multimodal input consisting of a numerical EEG representation, a waveform-based visual EEG representation, and the instruction text. 
For our method, we utilize Qwen2.5-VL-7B\cite{bai2025qwen25vl} as our LLM backbone with the native visual encoder, and EEGPT\cite{wang2024eegpt} as the numerical EEG signal encoder.
The signal stream encodes numerical EEG representations with a pretrained EEG encoder, and projects the resulting features into the LLM hidden space through MoP. The visual stream processes the waveform image through the native visual pathway of the MLLM, and the text stream follows the standard tokenization and embedding procedure.
The resulting EEG, visual, and text tokens are concatenated and fed into the language model for answer generation. 

\subsection{Dual-stream EEG Input Modeling}


{ 
We first describe how OneEEG handles channel-compatible and non-compatible EEG datasets. We then introduce how numerical and visual EEG representations are encoded and organized for the LLM backbone.
}

\subsubsection{Channel-Compatible Input Modes}
\label{sec:input_modes}

{
Following the dataset taxonomy in Sec.~\ref{taxonomy}, OneEEG adopts two input modes for both task-oriented QA and basic EEG attribute QA according to the channel setting. 
For standard-setting EEG frames, OneEEG uses both numerical and visual EEG representations. 
For non-standard-setting EEG frames, OneEEG omits the numerical EEG signal representation and uses only the visual EEG representation.
}

\subsubsection{Numerical Stream}

OneEEG uses the numerical signal when it is compatible with the selected 19-channel setting.
Given an EEG frame with $C$ channels and $T$ sampled time points, we denote the numerical signal as
\[
\mathbf{X} \in \mathbb{R}^{C \times T}.
\]
{Subsequently, the EEG numerical signal encoder partitions $\mathbf{X}$ along the temporal dimension into non-overlapping patches. 
With a patch length of $P$, the input is divided into $N_e=T/P$ temporal patches:
\[
\mathbf{X}
=
\{\mathbf{X}_i\}_{i=1}^{N_e},
\quad
\mathbf{X}_i \in \mathbb{R}^{C \times P}.
\]
Each patch contains the multi-channel EEG signals within a short time interval. 
The encoder encodes these temporal patches with channel-aware embeddings and Transformer layers, producing a sequence of EEG token features:
\[
\mathbf{Z}_{\mathrm{eeg}}
=
f_{\mathrm{eeg}}(\mathbf{X})
\in
\mathbb{R}^{N_e \times d_e},
\]
where $N_e$ denotes the number of EEG tokens and $d_e$ is the EEG encoder feature dimension.}

\begin{figure*}
    \centering
    \includegraphics[width=1.\linewidth]{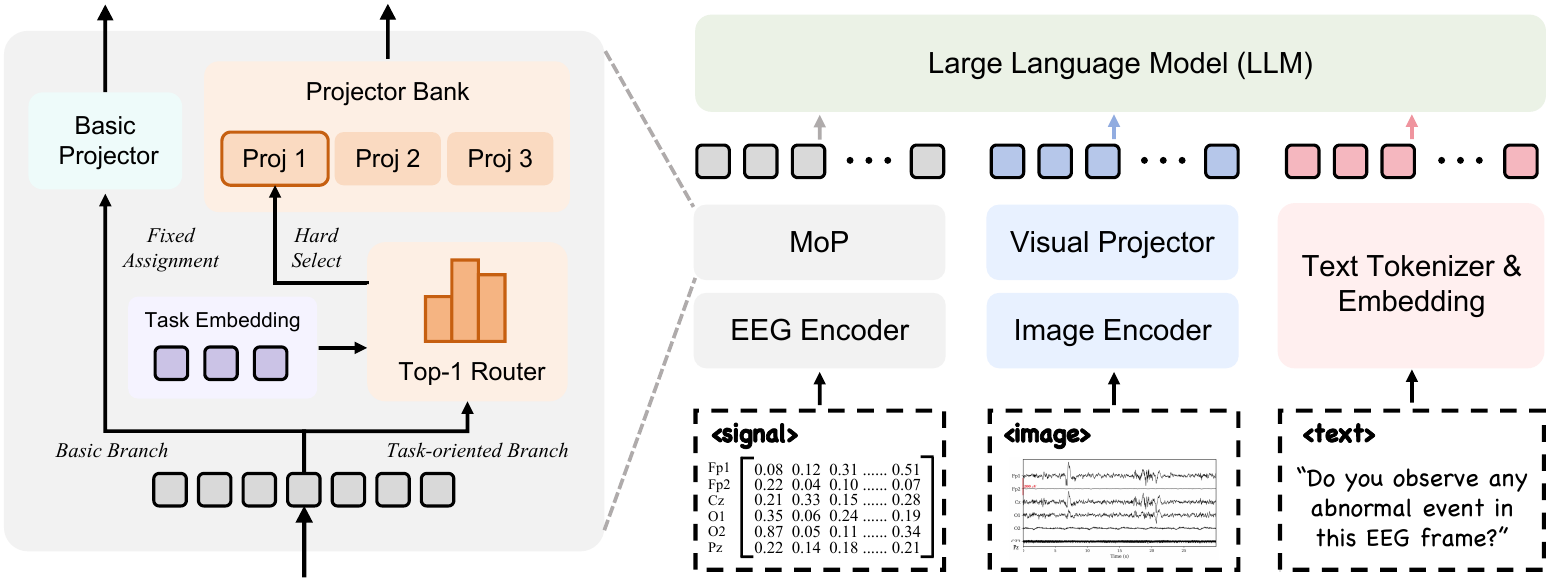}
    \caption{
    Overview of OneEEG. OneEEG integrates numerical EEG representations, visual EEG representations, and text instructions into a unified MLLM input. 
    {OneEEG uses both numerical and visual streams for standard-setting datasets and the visual stream only for non-standard-setting datasets.}
    The left panel details MoP: basic EEG attribute questions are assigned to a fixed basic projector, while task-oriented questions are handled by a router branch that selects one projector from the projector bank.
    }
    \label{fig:framework}
\end{figure*}

The EEG token features are then mapped to the hidden dimension of the language model by MoP:
\[
\mathbf{H}_{\mathrm{eeg}}
=
\mathrm{MoP}(\mathbf{Z}_{\mathrm{eeg}}, t),
\quad
\mathbf{H}_{\mathrm{eeg}} \in \mathbb{R}^{N_e \times d_l},
\]
where $d_l$ is the language-model hidden dimension and $t$ denotes the task-family label used by MoP. 

\subsubsection{Visual Stream}


For the visual stream, OneEEG takes the visual EEG representation, i.e., the waveform image $\mathbf{I}$, as input and processes it through the native visual module of the underlying Qwen2.5-VL backbone:
\[
\mathbf{H}_{\mathrm{vis}}
=
P_v(f_v(\mathbf{I}))
\in
\mathbb{R}^{N_v \times d_l},
\]
where $f_v$ denotes the pretrained visual encoder, $P_v$ denotes the original visual projection module, $N_v$ is the number of visual tokens, and $d_l$ is the hidden dimension of the language model.

\subsubsection{Prefix-Augmented Visual Conditioning}

For channel-compatible EEG frames, OneEEG further uses prefix-augmented visual conditioning to place waveform-derived visual tokens near the EEG tokens. 
In prefix-augmented visual conditioning, the waveform image is used in two ways. 
First, its visual features are appended after the EEG tokens as an auxiliary visual prefix:
\[
\mathbf{H}_{\mathrm{prefix}}
=
[\mathbf{H}_{\mathrm{eeg}};\mathbf{H}_{\mathrm{vis}}^{p}],
\]
where $\mathbf{H}_{\mathrm{vis}}^{p}$ denotes the waveform-image features used as prefix tokens.

Second, the original image placeholder is retained in the text prompt, allowing the same waveform image to be processed through the native image pathway of the MLLM. 
The final input sequence is written as
\[
\mathbf{H}_{\mathrm{input}}
=
[\mathbf{H}_{\mathrm{prefix}};\mathbf{H}_{\mathrm{prompt}}(\mathbf{I}, q)],
\]
where $\mathbf{H}_{\mathrm{prompt}}(\mathbf{I}, q)$ denotes the standard MLLM prompt representation formed by the waveform image $\mathbf{I}$ and the text instruction $q$.

The two visual paths serve different roles. The native image pathway preserves the pretrained image-text interface of the MLLM, whereas the auxiliary prefix provides local visual context adjacent to the newly introduced EEG signal tokens. This design helps bridge EEG signal features with waveform patterns in the pretrained visual space.

For non-standard channel settings where signal-side EEG tokens are omitted, OneEEG relies on the native waveform-image pathway without applying prefix-augmented visual conditioning. In this case, the input sequence reduces to the native MLLM prompt representation:
\[
\mathbf{H}_{\mathrm{input}}
=
\mathbf{H}_{\mathrm{prompt}}(\mathbf{I}, q),
\]
where $\mathbf{H}_{\mathrm{prompt}}(\mathbf{I}, q)$ contains the native waveform-image tokens and the text-instruction tokens.

\subsection{Task-aware Mixture of Projectors}

Given EEG features from the frozen EEG numerical signal encoder, OneEEG uses a task-aware \textbf{Mixture of Projectors (MoP)} module to map them into the LLM hidden space. 
As shown in Fig.~\ref{fig:framework}, instead of using a single shared EEG-to-LLM projector, MoP maintains a lightweight projector bank and selects the projection pathway according to the EEG token features and the task embedding.
Specifically, basic EEG attribute questions are assigned to a fixed individual projector, while task-oriented questions are routed to candidate projectors through a hard-selected router. 

\subsubsection{Projector Bank}

MoP maintains a lightweight EEG-to-LLM projector bank:
\[
\mathcal{P}
=
\{P_1, P_2, \ldots, P_K, P_{\mathrm{basic}}\},
\]
where $P_1,\ldots,P_K$ are candidate projectors for task-oriented questions and $P_{\mathrm{basic}}$ is the fixed projector for basic EEG attribute questions. Each projector is implemented as an MLP with the same architecture and maps EEG encoder features to the LLM hidden dimension:
\[
P_k: \mathbb{R}^{d_e} \rightarrow \mathbb{R}^{d_l},
\]
where $d_e$ and $d_l$ denote the EEG encoder feature dimension and the LLM hidden dimension, respectively.

For task-agnostic basic EEG attribute questions, OneEEG bypasses the router and directly applies the basic projector:
\[
\mathbf{H}_{\mathrm{eeg}}
=
P_{\mathrm{basic}}(\mathbf{Z}_{\mathrm{eeg}}),
\quad q \in \mathcal{Q}_{\mathrm{basic}}.
\]
For task-oriented questions, the projection path is selected from $\{P_1,\ldots,P_K\}$ by the task-oriented router.

\subsubsection{Task-oriented Router}

For task-oriented EEG questions, OneEEG uses a router to select one projection path from the projector bank. 
Unlike basic EEG attribute questions that follow the basic projector, task-oriented samples are routed among candidate projector slots according to both EEG token features and the task embedding. 
The task embedding is derived from the task family of each sample and provides task context for projector selection.

Given EEG encoder features
$\mathbf{Z}_{\mathrm{eeg}}=\{\mathbf{z}_i\}_{i=1}^{N_e}$ and the task-family label $t$, OneEEG obtains the corresponding task embedding $\mathbf{e}_t$.
The router concatenates this task embedding with each EEG token and computes token-level routing scores:
\[
\mathbf{a}_i
=
W_g [\mathbf{z}_i;\mathbf{e}_t],
\quad
\mathbf{a}_i \in \mathbb{R}^{K}.
\]
where $K$ is the number of candidate projector slots. 
The token-level scores are averaged across EEG tokens to obtain sequence-level routing scores:
\[
\mathbf{s}
=
\frac{1}{N_e}
\sum_{i=1}^{N_e}
\mathbf{a}_i .
\]
The selected projector is determined by hard top-1 routing:
\[
k^\ast
=
\arg\max_{k\in\{1,\ldots,K\}} s_k,
\quad
P_{k^\ast}\in\{P_1,\ldots,P_K\}.
\]
The EEG tokens are then mapped into the LLM hidden space by the selected projector:
\[
\mathbf{H}_{\mathrm{eeg}}
=
P_{k^\ast}(\mathbf{Z}_{\mathrm{eeg}}),
\quad q \in \mathcal{Q}_{\mathrm{task}}.
\]

We adopt a sparse routing strategy that activates only one projector slot for each task-oriented sample, keeping the EEG-to-LLM projection efficient without dense mixtures over all projectors.

\begin{figure}[t]
    \centering
    \includegraphics[width=0.85\linewidth]{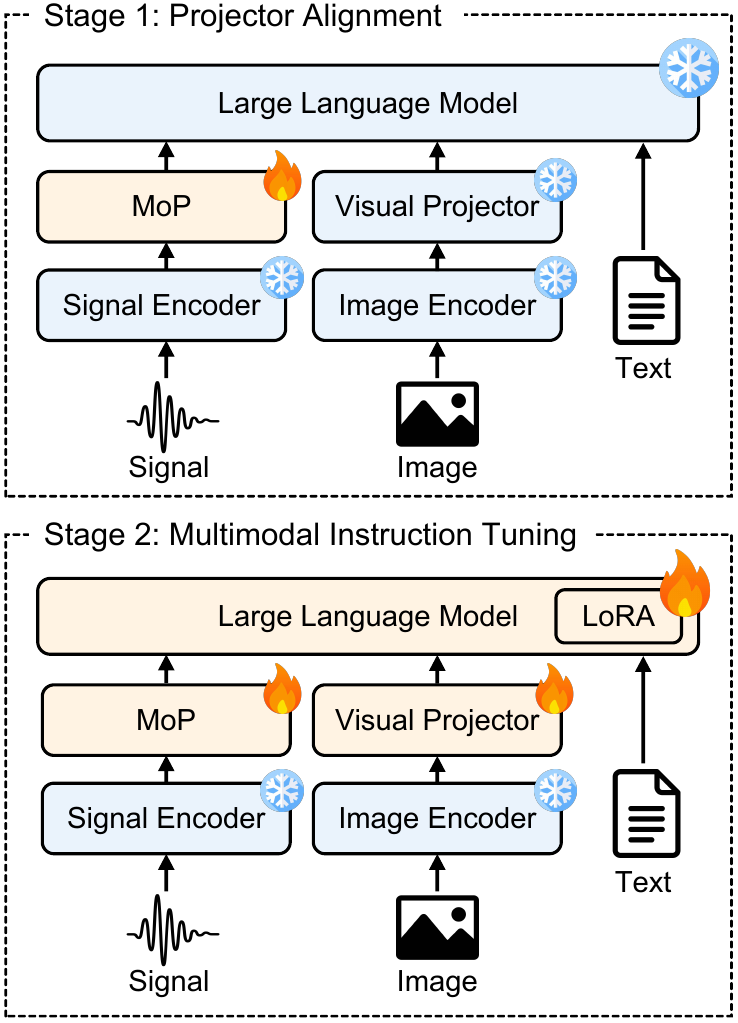}
    \caption{
    Two-stage training strategy of OneEEG. 
    In Stage 1, projector alignment trains MoP while keeping the signal encoder, image encoder, visual projector, and the LLM backbone frozen. 
    In Stage 2, multimodal instruction tuning further trains MoP, the visual projector, and LoRA parameters of the LLM, while the signal and image encoders remain frozen. 
    Snowflakes denote frozen modules, and flames denote trainable modules.} 
    \label{fig:stages}
\end{figure}

\subsection{Training Objectives and Stages}

OneEEG is optimized with a standard autoregressive language modeling objective and trained in two stages. The details are as follows.

\subsubsection{Training Objective}

OneEEG is trained with the standard autoregressive language modeling objective. Given the multimodal input embeddings $\mathbf{H}_{\mathrm{input}}$ and the target answer sequence $\mathbf{Y}=\{y_t\}_{t=1}^{M}$, the training loss is defined as
\[
\mathcal{L}_{\mathrm{LLM}}
=
-\sum_{t=1}^{M}
\log p_{\theta}
\left(
y_t \mid y_{<t}, \mathbf{H}_{\mathrm{input}}
\right).
\]

\subsubsection{Two-stage Training}
\label{sec:training_stages}

As shown in Fig.~\ref{fig:stages}, OneEEG is trained with a two-stage strategy, including projector alignment and multimodal instruction tuning.

In Stage 1, \textit{projector alignment}, the EEG encoder, visual module, and LLM backbone are frozen, and only the EEG-to-LLM projection modules are optimized. For MoP, this stage trains the projector bank under hard top-1 assignment, establishing an initial alignment between EEG encoder features and the LLM hidden space.

In Stage 2, \textit{multimodal instruction tuning}, OneEEG is initialized from the Stage 1 checkpoint. The EEG encoder and visual encoder remain frozen, while the MoP, the visual projector, and the LoRA\cite{hu2022lora} parameters are updated. This stage further adapts the alignment modules and instruction-following components while preserving the pretrained encoders and LLM backbone.

\section{Experiments}

In this section, we evaluate current MLLMs and OneEEG on our OneEEG-Bench and analyze the effectiveness of our method. Furthermore, we conduct detailed ablation experiments and in-depth analysis.

\begin{table*}[t]
\centering
\caption{
Main evaluation results on OneEEG-Bench. 
Invalid denotes the invalid-output rate, while all remaining columns report accuracy.}
\label{tab:main_results}
\setlength{\tabcolsep}{4.6pt}
\renewcommand{\arraystretch}{1.08}
\small
\begin{tabular}{@{}lcccccccccc@{}}
\toprule
\multirow{2}{*}{\textbf{Model}} &
  \multirow{2}{*}{\textbf{Invalid$\downarrow$}} &
  \multirow{2}{*}{\textbf{Basic}} &
  \multicolumn{4}{c}{\textit{Standard channel setting}} &
  \multicolumn{2}{c}{\textit{Non-standard channel setting}} &
  \multirow{2}{*}{\textbf{Overall}} &
  \multirow{2}{*}{\textbf{Macro}} \\
\cmidrule(lr){4-7} \cmidrule(lr){8-9}
& & & \textbf{Motor} & \textbf{State} & \textbf{Clinical} & \textbf{Event}
& \makebox[1.65cm][c]{\textbf{Sleep}} & \makebox[1.65cm][c]{\textbf{Vigilance}}
& & \\
\midrule
\rowcolor{gray!15}\multicolumn{11}{c}{\textit{Closed-source MLLMs}} \\
\midrule
Gemini-3 Flash \cite{gemini3flash} & 5.94 & 26.77 & 41.09 & 52.50 & 51.50 & 52.72 & 53.66 & 52.97 & 47.02 & 47.32 \\
Gemini-3 Pro \cite{gemini3pro} & 2.62 & 38.75 & 31.72 & 50.67 & 51.83 & 48.04 & 71.25 & 69.73 & 49.27 & 51.71 \\
GPT-5.5 \cite{gpt55} & 0.02 & 46.31 & 36.56 & 49.67 & 52.83 & 51.30 & 58.71 & 75.68 & 50.39 & 53.01 \\
\midrule
\rowcolor{gray!15}\multicolumn{11}{c}{\textit{Open-source MLLMs}} \\
\midrule
Qwen2.5-VL-7B \cite{bai2025qwen25vl} & 0.00 & 29.39 & 32.19 & 52.00 & 50.33 & 25.43 & 65.33 & 84.32 & 42.73 & 48.43 \\
Qwen3-VL-8B \cite{bai2025qwen3vl} & 0.00 & 32.18 & 33.59 & 52.00 & 48.50 & 29.78 & 44.43 & 45.41 & 39.41 & 40.84 \\
InternVL3.5-8B \cite{wang2025internvl35} & 0.00 & 27.26 & 37.03 & 48.50 & 49.50 & 30.54 & 62.72 & 82.16 & 43.22 & 48.24 \\
InternVL3.5-38B \cite{wang2025internvl35} & 0.00 & 29.23 & 32.34 & 50.50 & 49.50 & 30.65 & 64.98 & 85.41 & 43.56 & 48.94 \\
LLaVA-OneVision-2-8B \cite{an2026llavaonevision2} & 0.07 & 26.60 & 41.56 & 50.67 & 49.50 & 31.85 & 53.31 & 51.35 & 41.74 & 43.55 \\
\midrule
\rowcolor{gray!15}\multicolumn{11}{c}{\textit{Medical MLLMs}} \\
\midrule
MedGemma-1.5-4B \cite{sellergren2026medgemma} & 4.43 & 28.41 & 31.41 & 45.00 & 50.33 & 25.33 & 63.76 & 80.54 & 41.04 & 46.40 \\
Lingshu-7B \cite{xu2025lingshu} & 0.07 & 29.72 & 42.03 & 52.00 & 51.17 & 38.48 & 54.36 & 81.08 & 45.66 & 49.83 \\
Lingshu-32B \cite{xu2025lingshu} & 0.00 & 28.24 & 34.69 & 52.00 & 52.50 & 40.33 & 64.11 & 84.86 & 46.44 & 50.96 \\
\midrule
\rowcolor{gray!15}\multicolumn{11}{c}{\textit{EEG-specialized Models}} \\
\midrule
CerebraGloss \cite{gu2025cerebragloss} & 4.70 & 49.26 & 24.38 & 50.83 & 49.00 & 36.09 & 61.50 & 85.95 & 46.00 & 51.00 \\
\textbf{OneEEG} & 0.05 & \textbf{72.74} & \textbf{46.09} & \textbf{56.83} & \textbf{66.67} & \textbf{67.72} & \textbf{76.83} & \textbf{90.81} & \textbf{65.67} & \textbf{68.24} \\
\bottomrule
\end{tabular}
\end{table*}


\noindent\textbf{Benchmark and input setting.}
We evaluate all models on OneEEG-Bench, which contains basic EEG attribute QA and task-oriented EEG QA. 
Following the dataset taxonomy in Sec.~\ref{taxonomy}, we report results under standard-setting and non-standard-setting groups. 
In the main evaluation, OneEEG follows the channel-compatible input setting described in Sec.~\ref{sec:input_modes}. 
All baseline MLLMs are evaluated with the rendered EEG waveform image and textual question only, since these baselines do not support numerical EEG representations.

\noindent\textbf{Compared methods.}
We compare OneEEG with representative closed-source, open-source, medical, and EEG-specialized MLLMs, as listed in Table~\ref{tab:main_results}.

\noindent\textbf{Evaluation metrics.}
Each model output is normalized and matched against the candidate labels defined for the corresponding question. A prediction is counted as correct if it matches the ground-truth label and is marked as invalid if it cannot be mapped to any candidate label. For each task family, we report the overall accuracy over all samples belonging to that family, while \textit{Overall} applies the same accuracy measure to all evaluation samples in OneEEG-Bench. 
\textit{Macro} is the unweighted average of the seven family-level overall accuracies, including Basic, Motor, State, Clinical, Event, Sleep, and Vigilance. 
\textit{Invalid} is the proportion of outputs that cannot be mapped to any candidate label.


\subsection{Main Results}

Table~\ref{tab:main_results} reports the main results on OneEEG-Bench. 
Our proposed OneEEG achieves the best performance in both Overall and Macro scores, reaching 65.67\% and 68.24\%, respectively. 
Against the strongest baseline, GPT-5.5, OneEEG improves Overall accuracy by 15.28 and Macro accuracy by 15.23.
Compared with the Qwen2.5-VL backbone used by OneEEG, OneEEG brings substantial gains on Basic, Clinical, Event, Sleep, and Vigilance tasks, suggesting that EEG-specific training and signal-aware modeling are especially effective for clinically relevant EEG patterns and low-level physiological attributes, while gains on the Motor and State tasks are relatively smaller. 
In contrast to physiological attributes and clinical/event-related tasks, motor imagery and state-related EEG tasks require distinguishing more fine-grained EEG patterns, which still remain challenging.
These results show that OneEEG provides a stronger unified solution for multi-task EEG question answering, while also indicating room for improving signal representations on fine-grained motor and state tasks.

\begin{figure}[t]
    \centering
    \includegraphics[width=0.98\linewidth]{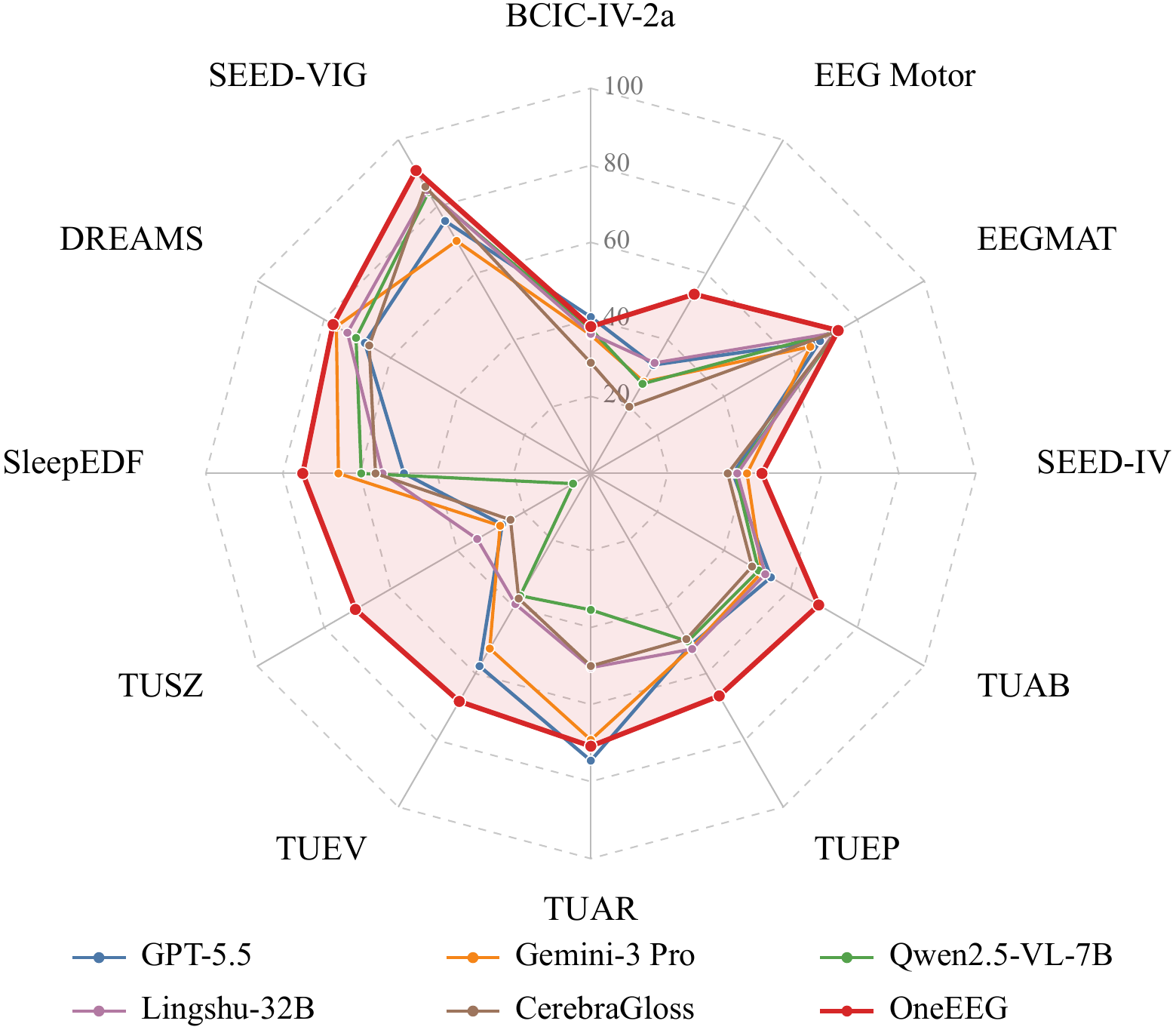}
    \caption{
    Per-dataset evaluation results on OneEEG-Bench. 
    Each axis reports the accuracy computed over all samples from the corresponding dataset, with larger values indicating better performance. 
    For readability, only representative baselines are shown in the radar plot.
    }
    \label{fig:perdataset}
\end{figure}

The baseline results further reveal the difficulty of OneEEG-Bench. 
Closed-source models generally perform better than open-source models. For instance, GPT-5.5 reaches 50.39\% Overall accuracy, while the best open-source baseline, InternVL3.5-38B, obtains 43.56\%. 
In addition, larger models tend to perform better: InternVL3.5-38B outperforms InternVL3.5-8B, and Lingshu-32B is better than Lingshu-7B, which indicates the benefit of model capacity in improving performance.
As a specialized EEG-trained baseline, CerebraGloss performs competitively on several tasks, but its training data mainly cover sleep and clinical datasets (HMC, DREAMS, and TUH-series datasets), which restricts its performance under the broader task and dataset coverage of OneEEG-Bench.

Fig.~\ref{fig:perdataset} further provides a dataset-level view of the results. 
OneEEG shows more consistent performance across EEG datasets, indicating that its advantage is not driven by a single dataset or task family. 
At the same time, the gains are not uniform across all datasets, suggesting that the frozen EEG encoder still provides imperfect and task-dependent signal features. 







\subsection{Ablation Study}

\begin{table*}[t]
\centering
\caption{
Ablation study on the core design choices of OneEEG under the Phase 1 setting. 
All metrics follow the same definitions as in the main evaluation, restricted to the Phase 1 evaluation set.
}
\label{tab:main_ablation}
\setlength{\tabcolsep}{4.6pt}
\renewcommand{\arraystretch}{1.08}
\small
\begin{tabular}{@{}lccc ccccc cc@{}}
\toprule
\multirow{2}{*}{\textbf{Variant}} &
  \multirow{2}{*}{\textbf{Projector}} &
  \multirow{2}{*}{\textbf{Route Strategy}} &
  \multirow{2}{*}{\textbf{Visual Prefix}} &
  \multicolumn{5}{c}{\textit{Task Family}} &
  \multirow{2}{*}{\textbf{Overall}} &
  \multirow{2}{*}{\textbf{Macro}} \\
\cmidrule(lr){5-9}
& & & & \textbf{Basic} & \textbf{Motor} & \textbf{State} & \textbf{Clinical} & \textbf{Event} & & \\
\midrule
Shared
& Shared  & -- & --
& 67.96 & \textbf{44.69} & 47.83 & 63.00 & 59.13 & 56.42 & 56.52 \\

Shared + Prefix
& Shared & -- & \checkmark
& 67.78 & 41.88 & 49.33 & 65.17 & \textbf{61.09} & 57.06 & 57.05 \\

MoP-Free
& MoP & Free & --
& 67.78 & 39.84 & 43.33 & 56.83 & 58.48 & 53.33 & 53.25 \\

MoP-Free + Prefix
& MoP & Free & \checkmark
& 69.63 & 43.91 & \underline{50.83} & \underline{65.50} & 58.59 & \underline{57.39} & \underline{57.69} \\

MoP
& MoP & Fixed & --
& \textbf{71.11} & 39.22 & 43.00 & 64.17 & 55.00 & 54.06 & 54.50 \\

OneEEG
& MoP & Fixed & \checkmark
& \underline{70.19} & \underline{44.53} & \textbf{53.33} & \textbf{69.17} & \underline{60.65} & \textbf{59.30} & \textbf{59.57} \\
\bottomrule
\end{tabular}
\end{table*}

\begin{table*}[!t]
\centering
\caption{
Ablation on EEG visual representations. 
We compare waveform images, spectrogram images, and their combination using GPT-5.5 and Qwen2.5-VL-7B. 
All metrics follow the same definitions as in the main evaluation.
}
\label{tab:wave_spec_ablation}
\setlength{\tabcolsep}{4.4pt}
\renewcommand{\arraystretch}{1.08}
\small
\begin{tabular}{@{}llccccccccc@{}}
\toprule
\multirow{2}{*}{\textbf{Model}} &
  \multirow{2}{*}{\textbf{Input}} &
  \multirow{2}{*}{\textbf{Basic}} &
  \multicolumn{4}{c}{\textit{Standard channel setting}} &
  \multicolumn{2}{c}{\textit{Non-standard channel setting}} &
  \multirow{2}{*}{\textbf{Overall}} &
  \multirow{2}{*}{\textbf{Macro}} \\
\cmidrule(lr){4-7} \cmidrule(lr){8-9}
& & & \textbf{Motor} & \textbf{State} & \textbf{Clinical} & \textbf{Event}
& \makebox[1.25cm][c]{\textbf{Sleep}} & \makebox[1.65cm][c]{\textbf{Vigilance}}
& & \\
\midrule
\multirow{3}{*}{GPT-5.5} & Waveform & \textbf{46.31} & 36.56 & 49.67 & \textbf{52.83} & \textbf{51.30} & 58.71 & \textbf{75.68} & \textbf{50.39} & \textbf{53.01} \\
 & Spectrogram & 28.41 & \textbf{36.72} & 49.33 & 50.00 & 32.83 & \textbf{60.10} & 55.68 & 42.49 & 44.72 \\
 & Waveform+Spectrogram & 45.65 & 33.44 & \textbf{50.67} & 52.00 & 50.65 & 55.57 & 75.14 & 49.22 & 51.87 \\
\midrule
\multirow{3}{*}{Qwen2.5-VL-7B} & Waveform & \textbf{29.39} & \textbf{32.19} & \textbf{52.00} & \textbf{50.33} & \textbf{25.43} & 65.33 & \textbf{84.32} & \textbf{42.73} & \textbf{48.43} \\
 & Spectrogram & 26.60 & 31.09 & 41.17 & 49.33 & 20.98 & 63.76 & 76.22 & 38.86 & 44.16 \\
 & Waveform+Spectrogram & 27.59 & 29.69 & 51.33 & 49.83 & 24.35 & \textbf{67.42} & 83.24 & 41.91 & 47.64 \\
\bottomrule
\end{tabular}
\end{table*}

\begin{figure*}[!t]
    \centering
    \includegraphics[width=1.\linewidth]{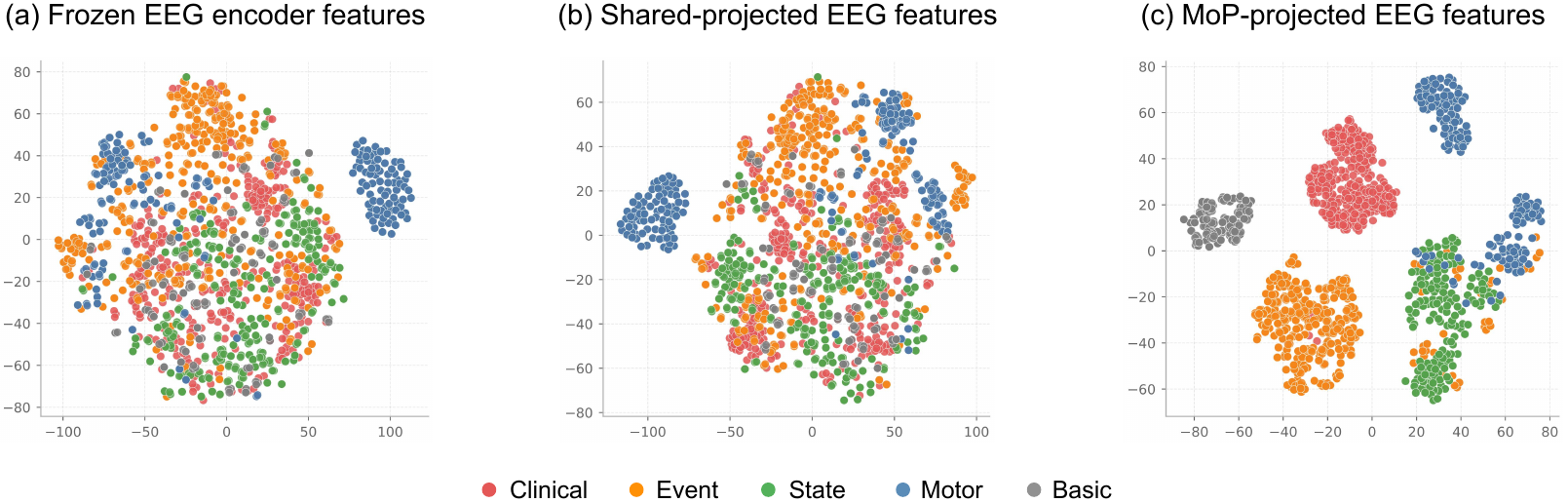}
    \caption{
    t-SNE visualization of EEG representations before projection (a) and after projection (b), (c). 
    Each point denotes an evaluation sample, and colors indicate task families. 
    MoP produces more structured family-level grouping after projection.
    }
    \label{fig:tsne}
\end{figure*}

Table~\ref{tab:main_ablation} studies the core design choices of OneEEG. 
We perform the ablation in Phase 1 to isolate the EEG-to-LLM projection design, since this stage freezes the LLM backbone and visual modules and only optimizes the EEG projection components, which allows a cleaner comparison of the projector structure, routing strategy, and prefix-augmented visual conditioning. 
We compare the full OneEEG with several variants, including a shared-projector baseline, a free-routing MoP variant where basic questions are routed together with task questions, and variants with or without visual prefix conditioning. 
In Table~\ref{tab:main_ablation}, ``Free'' denotes routing basic questions together with task questions, while ``Fixed'' assigns basic questions to the basic projector. 
Macro denotes the unweighted average over Basic, Motor, State, Clinical, and Event families.

The results show that MoP is the key component of OneEEG. 
With prefix-augmented visual conditioning, the full OneEEG achieves the best Overall and Macro scores, reaching 59.30\% and 59.57\%, respectively. 
It outperforms both Shared w/ Prefix and MoP-Free w/ Prefix in these two aggregate metrics, indicating that neither a single shared projector nor unconstrained routing can fully replace the task-aware MoP design.  
Prefix-augmented visual conditioning further improves MoP: adding an additional visual prefix to MoP improves overall accuracy from 54.06\% to 59.30\% and macro accuracy from 54.50\% to 59.57\%. 
However, the prefix alone is insufficient, as Shared w/ Prefix is still inferior to OneEEG in Overall and Macro. 
These results suggest that MoP serves as the primary mechanism for task-aware EEG-to-LLM projection, while the visual prefix provides useful context.

\subsection{In-Depth Analysis}
\label{sec:in_depth}

Beyond the main results and ablation study, we further conduct additional analyses to better understand the behavior of OneEEG, including the effect of visual EEG representations, feature-space visualization, routing-strategy comparison, and projector-usage analysis.

\begin{figure}[!t]
\centering
        \includegraphics[width=\linewidth]{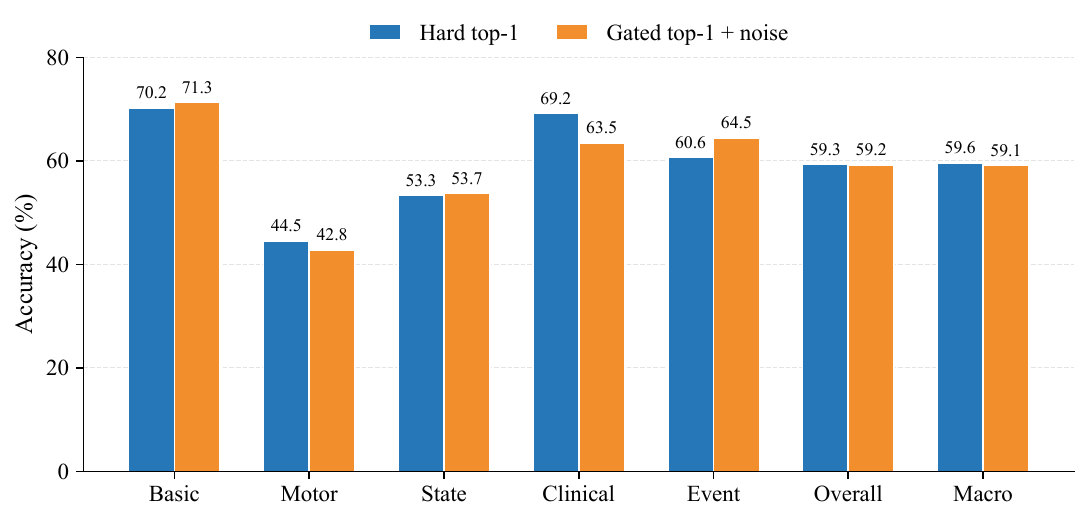}
        \caption{
        Comparison of MoP router variants under the Phase 1 setting.
        Hard top-1 routing and gated top-1 routing with noise achieve comparable Overall and Macro performance.
        }
        \label{fig:hard_soft}
\end{figure}

\subsubsection{Effect of Visual Representations}
\label{sec:visual_representations}
We first analyze how different EEG visual representations affect the performance of general MLLMs. 
As shown in Table~\ref{tab:wave_spec_ablation}, we compare waveform images, spectrogram images, and their combination using GPT-5.5 and Qwen2.5-VL-7B without additional training. 
Waveform images provide the most stable performance across the two models. 
For GPT-5.5, waveform input achieves the best Overall and Macro scores, reaching 50.39\% and 53.01\%, respectively. 
For Qwen2.5-VL-7B, waveform input also obtains the best Overall and Macro scores, with 42.73\% and 48.43\%, respectively. 
In contrast, spectrogram-only input leads to clear performance degradation, especially on Vigilance, where GPT-5.5 drops from 75.68\% to 55.68\%, and Qwen2.5-VL-7B drops from 84.32\% to 76.22\%. 
Combining waveform and spectrogram does not consistently improve over waveform-only input. 
These results suggest that waveform images are the most reliable visual EEG representation for current MLLMs, and we therefore use waveform images as the default visual input in our training process.

\subsubsection{Feature Visualization}
\label{sec:feature_visualization}
We further visualize the EEG features before and after the EEG-to-LLM projection using t-SNE. 
As shown in Fig.~\ref{fig:tsne}, the frozen EEG encoder features before projection are highly mixed across task families, indicating that the pretrained EEG encoder alone does not directly organize features according to the downstream QA taxonomy. 
The shared-projector baseline slightly reshapes the feature space, but different task families remain largely entangled. 
In contrast, MoP-projected features exhibit clearer family-level structures, with Basic, Clinical, Motor, State, and Event samples forming more distinguishable regions. 
This visualization suggests that MoP helps organize EEG features into a more task-aware projected space for the LLM. 
The remaining overlap between some families also indicates that fine-grained EEG representation alignment remains challenging.

\subsubsection{Effect of Routing Strategy}
\label{sec:routing_strategy}
We further compare two routing strategies of MoP under the Phase 1 setting. 
Besides the hard top-1 router used in the main configuration, we also evaluate a gated top-1 router with routing noise~\cite{shazeer2017routing_noise}, which follows a commonly used soft gating design for encouraging routing exploration during training. 
As shown in Fig.~\ref{fig:hard_soft}, the two variants achieve very similar aggregate performance. 
Hard top-1 routing obtains 59.30\% Overall and 59.57\% Macro, while gated top-1 routing with noise obtains 59.24\% Overall and 59.15\% Macro, only 0.06 and 0.42 percentage points lower, respectively.
This result suggests that the gain of MoP mainly comes from the multi-projector formulation rather than a specific router realization. 
By assigning different EEG task families to separate projection paths, MoP reduces the interference that may arise when heterogeneous EEG features are forced to share a single EEG-to-LLM projector. 
The comparable performance of the two router variants further indicates that sparse task-aware projection is a robust design choice for EEG-to-LLM alignment.
\subsubsection{Router Projector Usage}
\label{sec:router_usage}
\begin{figure}[!t]
\centering
        \includegraphics[width=\linewidth]{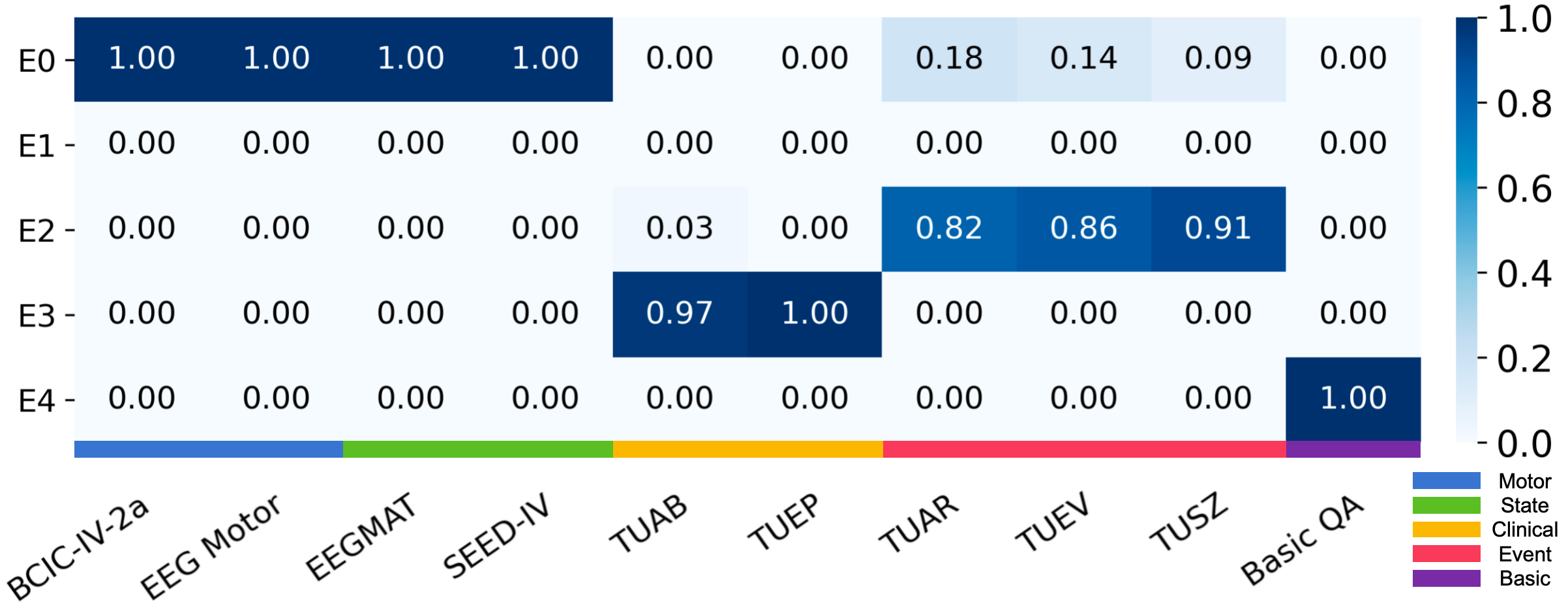}
        \caption{
        MoP router projector usage under hard top-1 routing.
        Rows denote projector slots and columns denote datasets, with each cell showing the selection ratio.
        }
        \label{fig:heatmap}
\end{figure}

We further analyze the projector usage induced by the hard routing configuration of MoP. 
Since each sample activates one projection path, the usage map in Fig.~\ref{fig:heatmap} provides a direct view of how samples are mapped to projector slots. 
The assignments show clear task-family preferences rather than uniform usage. 
Specifically, Clinical and Event tasks show relatively stable and distinct projector usage, while Motor and State datasets are more likely to share projector slots. 
This suggests that Clinical and Event tasks are more clearly separated in the current routing space, whereas Motor and State remain harder to distinguish. 
The observation is also consistent with the t-SNE visualization, where Motor and State representations remain partially overlapped. 
Overall, the usage pattern supports the interpretability of hard routing and indicates that more discriminative signal representations may further improve motor- and state-related EEG understanding.

\section{Conclusion}

In this work, we proposed OneEEG, a unified EEG-MLLM framework that jointly models numerical EEG representations, visual EEG representations, and text instructions for multi-task EEG question answering. 
OneEEG introduces a MoP module to align heterogeneous EEG signal features with the LLM hidden space. 
To support training and evaluation, we constructed OneEEG-Instruct and OneEEG-Bench, covering diverse EEG tasks, channel settings, and closed-set QA formats. 
Extensive experiments show that OneEEG consistently outperforms general, medical, and EEG-specialized MLLM baselines across heterogeneous EEG tasks. 
Further analyses demonstrate the effectiveness of our proposed OneEEG.

\bibliographystyle{IEEEtran}

\bibliography{references}

@article{qiu2026eegvlm,
  title={EEG-VLM: A Hierarchical Vision-Language Model With Multi-Level Feature Alignment and Visually Enhanced Language-Guided Reasoning for EEG Image-Based Sleep Stage Prediction},
  author={Qiu, Xihe and Ma, Gengchen and Wang, Haoyu and Zhan, Chen and Tan, Xiaoyu and Li, Shuo},
  journal={IEEE Journal of Biomedical and Health Informatics},
  year={2026},
  publisher={IEEE}
}

@article{zeng2025wavemind,
  title={Wavemind: Towards a conversational eeg foundation model aligned to textual and visual modalities},
  author={Zeng, Ziyi and Cai, Zhenyang and Cai, Yixi and Wang, Xidong and Chen, Junying and Wang, Rongsheng and Liu, Yipeng and Cai, Siqi and Wang, Benyou and Zhang, Zhiguo and others},
  journal={arXiv preprint arXiv:2510.00032},
  year={2025}
}

@inproceedings{gu2025cerebragloss,
  title={CerebraGloss: Instruction-Tuning a Large Vision-Language Model for Fine-Grained Clinical EEG Interpretation},
  author={Gu, Wei and Luo, Tianming and Zhang, Qiran and Ye, Mohan and Shen, Xiao and Chen, Wenxin and Li, Yunhuan and Zhang, Yichen and Hong, Jing and Lu, Bao-liang and Zheng, Wei-Long},
  booktitle={The Fourteenth International Conference on Learning Representations},
  year={2026}
}

@article{ma20262e2llm,
  title={E\^{} 2-LLM: Bridging Neural Signals and Interpretable Affective Analysis},
  author={Ma, Fei and Lin, Han and Xie, Yifan and Ren, Hongwei and Shen, Xiaoyu and Ding, Wenbo and Tian, Qi},
  journal={arXiv preprint arXiv:2601.07877},
  year={2026}
}

@article{wang2024eegpt,
  title={Eegpt: Pretrained transformer for universal and reliable representation of eeg signals},
  author={Wang, Guangyu and Liu, Wenchao and He, Yuhong and Xu, Cong and Ma, Lin and Li, Haifeng},
  journal={Advances in Neural Information Processing Systems},
  volume={37},
  pages={39249--39280},
  year={2024}
}

@inproceedings{jiang2024labram,
  title={Large brain model for learning generic representations with tremendous EEG data in BCI},
  author={Jiang, Wei-Bang and Zhao, Liming and Lu, Bao-Liang},
  booktitle={International Conference on Learning Representations},
  volume={2024},
  pages={16405--16426},
  year={2024}
}

@article{jiang2024neurolm,
  title={NeuroLM: A universal multi-task foundation model for bridging the gap between language and EEG signals},
  author={Jiang, Wei-Bang and Wang, Yansen and Lu, Bao-Liang and Li, Dongsheng},
  journal={arXiv preprint arXiv:2409.00101},
  year={2024}
}

@article{zhang2023brant,
  title={Brant: Foundation model for intracranial neural signal},
  author={Zhang, Daoze and Yuan, Zhizhang and Yang, Yang and Chen, Junru and Wang, Jingjing and Li, Yafeng},
  journal={Advances in Neural Information Processing Systems},
  volume={36},
  pages={26304--26321},
  year={2023}
}

@inproceedings{wang2025cbramod,
  title={Cbramod: A criss-cross brain foundation model for eeg decoding},
  author={Wang, Jiquan and Zhao, Sha and Luo, Zhiling and Zhou, Yangxuan and Jiang, Haiteng and Li, Shijian and Li, Tao and Pan, Gang},
  booktitle={International conference on learning representations},
  volume={2025},
  pages={75310--75346},
  year={2025}
}

@article{xiao2026brainomni,
  title={Brainomni: A brain foundation model for unified eeg and meg signals},
  author={Xiao, Qinfan and Cui, Ziyun and Zhang, Chi and Chen, Siqi and Wu, Wen and Thwaites, Andrew and Woolgar, Alexandra and Zhou, Bowen and Zhang, Chao},
  journal={Advances in Neural Information Processing Systems},
  volume={38},
  pages={41179--41212},
  year={2026}
}

@article{yang2023biot,
  title={Biot: Biosignal transformer for cross-data learning in the wild},
  author={Yang, Chaoqi and Westover, M and Sun, Jimeng},
  journal={Advances in Neural Information Processing Systems},
  volume={36},
  pages={78240--78260},
  year={2023}
}

@inproceedings{cui2024neurogpt,
  title={Neuro-gpt: Towards a foundation model for eeg},
  author={Cui, Wenhui and Jeong, Woojae and Th{\"o}lke, Philipp and Medani, Takfarinas and Jerbi, Karim and Joshi, Anand A and Leahy, Richard M},
  booktitle={2024 IEEE International Symposium on Biomedical Imaging (ISBI)},
  pages={1--5},
  year={2024},
  organization={IEEE}
}

@misc{bai2025qwen25vl,
      title={Qwen2.5-VL Technical Report}, 
      author={Shuai Bai and Keqin Chen and Xuejing Liu and Jialin Wang and Wenbin Ge and Sibo Song and Kai Dang and Peng Wang and Shijie Wang and Jun Tang and Humen Zhong and Yuanzhi Zhu and Mingkun Yang and Zhaohai Li and Jianqiang Wan and Pengfei Wang and Wei Ding and Zheren Fu and Yiheng Xu and Jiabo Ye and Xi Zhang and Tianbao Xie and Zesen Cheng and Hang Zhang and Zhibo Yang and Haiyang Xu and Junyang Lin},
      year={2025},
      eprint={2502.13923},
      archivePrefix={arXiv},
      primaryClass={cs.CV},
      url={https://arxiv.org/abs/2502.13923}, 
}

@article{bai2025qwen3vl,
  title={Qwen3-vl technical report},
  author={Bai, Shuai and Cai, Yuxuan and Chen, Ruizhe and Chen, Keqin and Chen, Xionghui and Cheng, Zesen and Deng, Lianghao and Ding, Wei and Gao, Chang and Ge, Chunjiang and others},
  journal={arXiv preprint arXiv:2511.21631},
  year={2025}
}

@misc{an2026llavaonevision2,
      title={LLaVA-OneVision-2: Towards Next-Generation Perceptual Intelligence}, 
      author={Xiang An and Yin Xie and Feilong Tang and Yunyao Yan and Huajie Tan and Didi Zhu and Changrui Chen and Xiuwei Zhao and Bin Qin and Kaicheng Yang and Yifei Shen and Yuanhan Zhang and Kaichen Zhang and Wenkang Zhang and Zheng Cheng and Nansen Zhang and Chunsheng Wu and Chunjiang Ge and Zimin Ran and Dehua Song and Chunyuan Li and Shikun Feng and Ming Hu and Zhangquan Chen and Junbo Niu and Bo Li and Ziyong Feng and Ziwei Liu and Zongyuan Ge and Jiankang Deng},
      year={2026},
      eprint={2605.25979},
      archivePrefix={arXiv},
      primaryClass={cs.CV},
      url={https://arxiv.org/abs/2605.25979}, 
}

@article{wang2025internvl35,
  title={Internvl3. 5: Advancing open-source multimodal models in versatility, reasoning, and efficiency},
  author={Wang, Weiyun and Gao, Zhangwei and Gu, Lixin and Pu, Hengjun and Cui, Long and Wei, Xingguang and Liu, Zhaoyang and Jing, Linglin and Ye, Shenglong and Shao, Jie and others},
  journal={arXiv preprint arXiv:2508.18265},
  year={2025}
}

@misc{gpt55,
  author       = {{OpenAI}},
  title        = {{GPT-5.5 Thinking}},
  year         = {2026},
  howpublished = {\url{https://chatgpt.com/}},
  note         = {Large language model, accessed May 26, 2026}
}

@misc{gemini3pro,
  author       = {{Google DeepMind}},
  title        = {{Gemini 3 Pro}},
  year         = {2026},
  howpublished = {\url{https://deepmind.google/models/gemini/pro/}},
  note         = {Large language model, accessed May 26, 2026}
}

@misc{gemini3flash,
  author       = {{Google DeepMind}},
  title        = {{Gemini 3 Flash}},
  year         = {2025},
  howpublished = {\url{https://deepmind.google/models/model-cards/}},
  note         = {Model card, accessed May 26, 2026}
}

@article{xu2025lingshu,
  title={Lingshu: A generalist foundation model for unified multimodal medical understanding and reasoning},
  author={Xu, Weiwen and Chan, Hou Pong and Li, Long and Aljunied, Mahani and Yuan, Ruifeng and Wang, Jianyu and Xiao, Chenghao and Chen, Guizhen and Liu, Chaoqun and Li, Zhaodonghui and others},
  journal={arXiv preprint arXiv:2506.07044},
  year={2025}
}

@article{sellergren2026medgemma,
  title={Medgemma 1.5 technical report},
  author={Sellergren, Andrew and Gao, Chufan and Mahvar, Fereshteh and Kohlberger, Timo and Jamil, Fayaz and Traverse, Madeleine and Tono, Alberto and Sadjad, Bashir and Yang, Lin and Lau, Charles and others},
  journal={arXiv preprint arXiv:2604.05081},
  year={2026}
}

@article{chen2024llavamole,
  title={Llava-mole: Sparse mixture of lora experts for mitigating data conflicts in instruction finetuning mllms},
  author={Chen, Shaoxiang and Jie, Zequn and Ma, Lin},
  journal={arXiv preprint arXiv:2401.16160},
  year={2024}
}

@misc{brunner2008bciciv2a,
  author       = {Brunner, Clemens and Leeb, Robert and M{\"u}ller-Putz, Gernot R. and Schl{\"o}gl, Alois and Pfurtscheller, Gert},
  title        = {{BCI Competition 2008 -- Graz Data Set A}},
  year         = {2008},
  howpublished = {\url{https://www.bbci.de/competition/iv/desc_2a.pdf}},
  note         = {{BCI Competition IV, Dataset 2a}}
}

@article{eegmotor,
  author = {Schalk, Gerwin},
  title = {{EEG Motor Movement/Imagery Dataset}},
  journal = {{PhysioNet}},
  year = {2009},
  month = sep,
  note = {Version 1.0.0},
  doi = {10.13026/C28G6P},
  url = {https://doi.org/10.13026/C28G6P}
}

@article{zyma2019eegmat,
  title={Electroencephalograms during mental arithmetic task performance},
  author={Zyma, Igor and Tukaev, Sergii and Seleznov, Ivan and Kiyono, Ken and Popov, Anton and Chernykh, Mariia and Shpenkov, Oleksii},
  journal={Data},
  volume={4},
  number={1},
  pages={14},
  year={2019},
  publisher={MDPI}
}

@ARTICLE{seediv, 
author={W. Zheng and W. Liu and Y. Lu and B. Lu and A. Cichocki}, 
journal={IEEE Transactions on Cybernetics}, 
title={EmotionMeter: A Multimodal Framework for Recognizing Human Emotions}, 
year={2018}, 
volume={}, 
number={}, 
pages={1-13}, 
doi={10.1109/TCYB.2018.2797176}, 
ISSN={2168-2267}, 
month={},}

@article{obeid2016tuh,
  author  = {Obeid, Iyad and Picone, Joseph},
  title   = {{The Temple University Hospital EEG Data Corpus}},
  journal = {Frontiers in Neuroscience},
  volume  = {10},
  pages   = {196},
  year    = {2016},
  doi     = {10.3389/fnins.2016.00196}
}

@article{shah2018tusz,
  title={The temple university hospital seizure detection corpus},
  author={Shah, Vinit and Von Weltin, Eva and Lopez, Silvia and McHugh, James Riley and Veloso, Lillian and Golmohammadi, Meysam and Obeid, Iyad and Picone, Joseph},
  journal={Frontiers in neuroinformatics},
  volume={12},
  pages={83},
  year={2018},
  publisher={Frontiers Media SA}
}

@inproceedings{hamid2020tuar,
  author    = {Hamid, Amina and Gagliano, Kyle and Rahman, Shazia and Tobochnik, Steven and Obeid, Iyad and Picone, Joseph},
  title     = {{The Temple University Artifact Corpus: An Annotated Corpus of EEG Artifacts}},
  booktitle = {Proceedings of the IEEE Signal Processing in Medicine and Biology Symposium},
  year      = {2020},
  url       = {https://isip.piconepress.com/conferences/ieee_spmb/2020/papers/p01_09.pdf}
}

@misc{tuh_web,
  author       = {{Neural Engineering Data Consortium}},
  title        = {{Temple University Hospital EEG Corpus}},
  year         = {2026},
  howpublished = {\url{https://isip.piconepress.com/projects/nedc/html/tuh_eeg/}},
  note         = {Accessed May 26, 2026}
}

@article{kemp2000sleepedf,
  title={Analysis of a sleep-dependent neuronal feedback loop: the slow-wave microcontinuity of the EEG},
  author={Kemp, Bob and Zwinderman, Aeilko H and Tuk, Bert and Kamphuisen, Hilbert AC and Oberye, Josefien JL},
  journal={IEEE Transactions on Biomedical Engineering},
  volume={47},
  number={9},
  pages={1185--1194},
  year={2000},
  publisher={IEEE}
}

@dataset{dreams,
  author       = {Stephanie Devuyst},
  title        = {The DREAMS Databases and Assessment Algorithm},
  month        = jan,
  year         = 2005,
  publisher    = {Zenodo},
  doi          = {10.5281/zenodo.2650142},
  url          = {https://doi.org/10.5281/zenodo.2650142},
}

@article{seedvig,
	author={Wei-Long Zheng and Bao-Liang Lu},
	title={A multimodal approach to estimating vigilance using EEG and forehead EOG},
	journal={Journal of Neural Engineering},
	volume={14},
	number={2},
	pages={026017},
	url={http://stacks.iop.org/1741-2552/14/i=2/a=026017},
	year={2017},
}

@article{mabrainmoe,
  title={BrainMoE: Towards Universal EEG Foundation Models with Channel-wise Mixture-of-Experts},
  author={Ma, Fan and Jiang, Mingyang and Qian, Lingfei and Gu, Zhiling and Xu, Hua}
}

@article{gao2026eegmoe,
  title={EEGMoE: A Domain-Decoupled Mixture-of-Experts Model for Self-Supervised EEG Representation Learning},
  author={Gao, Xuange and Wang, Danli and Zhao, Yanyan},
  journal={IEEE Transactions on Neural Networks and Learning Systems},
  year={2026},
  publisher={IEEE}
}

@inproceedings{wang2022adamix,
  title={Adamix: Mixture-of-adaptations for parameter-efficient model tuning},
  author={Wang, Yaqing and Agarwal, Sahaj and Mukherjee, Subhabrata and Liu, Xiaodong and Gao, Jing and Hassan, Ahmed and Gao, Jianfeng},
  booktitle={Proceedings of the 2022 Conference on Empirical Methods in Natural Language Processing},
  pages={5744--5760},
  year={2022}
}

@article{zoph2022stmoe,
  title={St-moe: Designing stable and transferable sparse expert models},
  author={Zoph, Barret and Bello, Irwan and Kumar, Sameer and Du, Nan and Huang, Yanping and Dean, Jeff and Shazeer, Noam and Fedus, William},
  journal={arXiv preprint arXiv:2202.08906},
  year={2022}
}

@article{gui2024eegmamba,
  title={EEGMamba: Bidirectional state space model with mixture of experts for EEG multi-task classification},
  author={Gui, Yiyu and Chen, MingZhi and Su, Yuqi and Luo, Guibo and Yang, Yuchao},
  journal={arXiv preprint arXiv:2407.20254},
  year={2024}
}

@article{acharya2018seizure_survey,
  title={Deep convolutional neural network for the automated detection and diagnosis of seizure using EEG signals},
  author={Acharya, U Rajendra and Oh, Shu Lih and Hagiwara, Yuki and Tan, Jen Hong and Adeli, Hojjat},
  journal={Computers in biology and medicine},
  volume={100},
  pages={270--278},
  year={2018},
  publisher={Elsevier}
}

@article{wang2025emotion_reco,
  title={Band-level adaptive fusion network for cross-subject EEG emotion recognition},
  author={Wang, Yilin and Zhang, Li and Zhang, Yan},
  journal={IEEE Transactions on Instrumentation and Measurement},
  year={2025},
  publisher={IEEE}
}

@article{liu2024msvtnet,
  title={MSVTNet: Multi-scale vision transformer neural network for EEG-based motor imagery decoding},
  author={Liu, Ke and Yang, Tao and Yu, Zhuliang and Yi, Weibo and Yu, Hong and Wang, Guoyin and Wu, Wei},
  journal={IEEE Journal of Biomedical and Health Informatics},
  volume={28},
  number={12},
  pages={7126--7137},
  year={2024},
  publisher={IEEE}
}

@article{ozdemir2021emotion_method_2,
  title={EEG-based emotion recognition with deep convolutional neural networks},
  author={Ozdemir, Mehmet Akif and Degirmenci, Murside and Izci, Elif and Akan, Aydin},
  journal={Biomedical Engineering/Biomedizinische Technik},
  volume={66},
  number={1},
  pages={43--57},
  year={2021},
  publisher={De Gruyter}
}

@article{liu2024graph,
  title={Graph neural networks in eeg-based emotion recognition: a survey},
  author={Liu, Chenyu and Deng, Yuqiu and Wu, Yihao and Yang, Ruizhi and Wang, Zhongruo and Zhang, Liangwei and Chen, Siyun and Zhang, Tianyi and Liu, Yang and Ding, Yi and others},
  journal={arXiv preprint arXiv:2402.01138},
  year={2024}
}

@article{lawhern2018eegnet,
  title={EEGNet: a compact convolutional neural network for EEG-based brain--computer interfaces},
  author={Lawhern, Vernon J and Solon, Amelia J and Waytowich, Nicholas R and Gordon, Stephen M and Hung, Chou P and Lance, Brent J},
  journal={Journal of neural engineering},
  volume={15},
  number={5},
  pages={056013},
  year={2018},
  publisher={iOP Publishing}
}

@article{lee2025bcireview,
  title={A review of hybrid EEG-based multimodal human--computer interfaces using deep learning: applications, advances, and challenges},
  author={Lee, Hyung-Tak and Shim, Miseon and Liu, Xianghong and Cheon, Hye-Ran and Kim, Sang-Gyu and Han, Chang-Hee and Hwang, Han-Jeong},
  journal={Biomedical Engineering Letters},
  volume={15},
  number={4},
  pages={587--618},
  year={2025},
  publisher={Springer}
}

@article{satapathy2025sleepreview,
  title={Deep Learning-Based Sleep Stage Classification: A Systematic Review},
  author={Satapathy, Santosh Kumar and Patel, Tapan and Modi, Madhuram and Davra, Harsh and Brahma, Biswajit and Tripathy, Santosh Kumar and Bhoi, Akash Kumar},
  journal={Journal of The Institution of Engineers (India): Series B},
  volume={106},
  number={6},
  pages={2087--2109},
  year={2025},
  publisher={Springer}
}

@article{fratangelo2025point, 
  title={Point-of-care electroencephalography in acute neurological care: A narrative review}, 
  author={Fratangelo, Roberto and Lolli, Francesco and Scarpino, Maenia and Grippo, Antonello}, 
  journal={Neurology International}, 
  volume={17}, 
  number={4}, 
  pages={48}, 
  year={2025}, 
  publisher={MDPI} 
}

@article{yedukondalu2025cognitive, 
  title={Cognitive load detection through EEG lead wise feature optimization and ensemble classification}, 
  author={Yedukondalu, Jammisetty and Sunkara, Kalyani and Radhika, Vankayalapati and Kondaveeti, Sivakrishna and Anumothu, Murali and Murali Krishna, Yadadavalli}, 
  journal={Scientific Reports}, 
  volume={15}, 
  number={1}, 
  pages={842}, 
  year={2025}, 
  publisher={Nature Publishing Group UK London} 
}

@inproceedings{
hu2022lora,
title={Lo{RA}: Low-Rank Adaptation of Large Language Models},
author={Edward J Hu and Yelong Shen and Phillip Wallis and Zeyuan Allen-Zhu and Yuanzhi Li and Shean Wang and Lu Wang and Weizhu Chen},
booktitle={International Conference on Learning Representations},
year={2022},
url={https://openreview.net/forum?id=nZeVKeeFYf9}
}

@article{klem19991020channel,
  title={The ten-twenty electrode system of the international federation. The international federation of clinical neurophysiology},
  author={Klem, George H},
  journal={Electroencephalogr. Clin. Neurophysiol. Suppl.},
  volume={52},
  pages={3--6},
  year={1999}
}

@article{yang2025qwen3,
  title={Qwen3 technical report},
  author={Yang, An and Li, Anfeng and Yang, Baosong and Zhang, Beichen and Hui, Binyuan and Zheng, Bo and Yu, Bowen and Gao, Chang and Huang, Chengen and Lv, Chenxu and others},
  journal={arXiv preprint arXiv:2505.09388},
  year={2025}
}

@article{shazeer2017routing_noise,
  title={Outrageously large neural networks: The sparsely-gated mixture-of-experts layer},
  author={Shazeer, Noam and Mirhoseini, Azalia and Maziarz, Krzysztof and Davis, Andy and Le, Quoc and Hinton, Geoffrey and Dean, Jeff},
  journal={arXiv preprint arXiv:1701.06538},
  year={2017}
}

@article{zhang2021enhancing,
  author={Zhang, Lianbo and Huang, Shaoli and Liu, Wei},
  title={Enhancing Mixture-of-Experts by Leveraging Attention for Fine-Grained Recognition},
  journal={IEEE Transactions on Multimedia},
  volume={24},
  pages={4409--4421},
  year={2022},
  doi={10.1109/TMM.2021.3117064}
}

@article{jin2024pgcn,
  author={Jin, Ming and Du, Changde and He, Huiguang and Cai, Ting and Li, Jinpeng},
  title={{PGCN}: Pyramidal Graph Convolutional Network for {EEG} Emotion Recognition},
  journal={IEEE Transactions on Multimedia},
  volume={26},
  pages={9070--9082},
  year={2024},
  doi={10.1109/TMM.2024.3385676}
}

@article{pi2026enhancing,
  author={Pi, Yao and Li, Ming and Zhang, Xianbin and Wu, Wei and Qi, Xuanhao and Yue, Jihui and Bian, Guibin and Wu, Wanqing},
  title={Enhancing Generalization for {EEG}-Based Emotion Recognition Models With the Personality Cross-Attention Mechanism},
  journal={IEEE Transactions on Multimedia},
  pages={1--14},
  year={2026},
  doi={10.1109/TMM.2026.3673551}
}

\end{document}